\documentclass{article}
\usepackage{spconf,amsmath,graphicx}
\usepackage[utf8]{inputenc} 
\usepackage[T1]{fontenc}    
\usepackage{hyperref}       
\usepackage{url}            
\usepackage{booktabs}       
\usepackage{amsfonts}       
\usepackage{nicefrac}       
\usepackage{microtype}      
\usepackage{graphicx}
\usepackage{mathtools}
\usepackage{adjustbox}
\graphicspath{{./Figs/}}

\title{Looking in the Right place for Anomalies: Explainable AI through Automatic Location Learning}
\name{\begin{tabular} {c} Satyananda Kashyap, Alexandros Karargyris, Joy Wu, Yaniv Gur, Arjun Sharma \\ Ken C. L. Wong, Mehdi Moradi, Tanveer Syeda-Mahmood\end{tabular}}
\address{IBM Research Almaden}

\begin{document}
%
\maketitle

\begin{abstract}
	Deep learning has now become the de facto approach to the recognition of anomalies in medical imaging. Their 'black box' way of classifying medical images into anomaly labels poses problems for their acceptance, particularly with clinicians. Current explainable AI methods offer justifications through visualizations such as heat maps but cannot guarantee that the network is focusing on the relevant image region fully containing the anomaly. In this paper we develop an approach to explainable AI in which the anomaly is assured to be overlapping the expected location when present. This is made possible by automatically extracting location-specific labels from textual reports and learning the association of expected locations to labels using a hybrid combination of Bi-Directional Long Short-Term Memory Recurrent Neural Networks (Bi-LSTM) and DenseNet-121. Use of this expected location to bias the subsequent attention-guided inference network based on ResNet101 results in the isolation of the anomaly at the expected location when present. The method is evaluated on a large chest X-ray dataset. 
\end{abstract}

\section{Introduction}
Deep learning is now the dominant paradigm in AI. Despite the large number of layers now being used, the networks are optimized based on a single objective function through basic mechanisms of back propagation. As a result, it has become increasingly difficult to explain if the network output is driven by the relevant image region. This particularly poses a problem in radiology reads, where the clinicians not only need to be convinced that the machine is classifying correctly, but that it is driven by features within the relevant region containing the anomaly \cite{nprarticle}. Unlike rule-based expert systems or decision tree methods which can externalize the feature choices, deep neural networks do not automatically provide explainable features. Current explainable AI methods focus on post explanations by either generating visual explanations such as layer-wise relevance propagation \cite{plos1} and heat maps, transforming the labels to show higher level semantic associations \cite{eccv16} or rationalizing neural predictions by identifying relevant regions in text [5]. The relevancy of regions is also built into the object detection formalisms in computer vision where current networks such as RetinaNet solve for both the object category and its predicted image location \cite{acl2016}. These formalisms typically search for objects through bounding boxes in several locations, and orientations through single pass \cite{retinanet,yolo2017} or two pass networks \cite{NIPS2015_5638} that first propose candidate region and then classify the regions to match in both location and label for the object. However, none of these approaches can guarantee that the network is focusing in the relevant image region. 

In this paper we develop an approach to explainable AI in which the anomaly is designed to be overlapping the expected location when present. This is made possible by automatically extracting location-specific labels from textual reports and learning the association of expected locations to labels using a Bi-Directional Long Short-Term Memory Recurrent Neural Networks (Bi-LSTM). Use of this expected location to bias the subsequent attention-guided inference network based on ResNet101 results in the isolation of the anomaly at the expected location when present. The method is evaluated on a large chest x-ray dataset. Comparison to direct image-based classification approaches shows an improvement in accuracy of classification while still ensuring that the anomaly is localized in its expected location, thus leading to a better justification for the learned model. Some of the large scale medical image collections are accompanied by radiology text reports. Mining these reports for possible location cues can generate location hints.  This approach was used recently to annotate for anomalies in earlier work \cite{moradi2018}. In \cite{moradi2018}, MESH terms extracted from text paragraphs were used in an LSTM formulation to predict the location. However, not all terms in the text may be relevant for the localization. While the unidirectional LSTM formulation can model the sequence order in dictated reports, due to the variations in which anomalies and their locations are described, missing and spurious terms can occur requiring the use of bi-directional models. Finally, the end point of the localization approaches is the regression of the location coordinates whereas the problem addressed in this paper is the classification of the anomalies taking the expected location into account for improved accuracy. The main contributions of the paper are as follows: 
\begin{enumerate}
	\item Automated inference of findings and their locations from reports using natural language analysis of text guided by clinical knowledge assembled through a chest X-ray lexicon.
	\item Use of a hybrid multimodal learning formulation for the localization of regions guided by a sequence of textual labels characterizing labels during training.
	\item Use of the automatically predicted locations to bias the classification using an attention-guided inference network based on ResNet101.
\end{enumerate}

The classification problem that we solve using the proposed approach is detection of laterality of an opacity finding in a chest X-ray image. In this scenario, the image is known to include opacity, but it is not clear if the right or the left lung is involved. The choice of the problem is driven by our effort to create AI solutions for interpretation of chest X-ray that go beyond a simple global label for the image.  

\section{Methods}

\begin{figure}[htbp]
	\begin{centering}
		\includegraphics[width=\columnwidth]{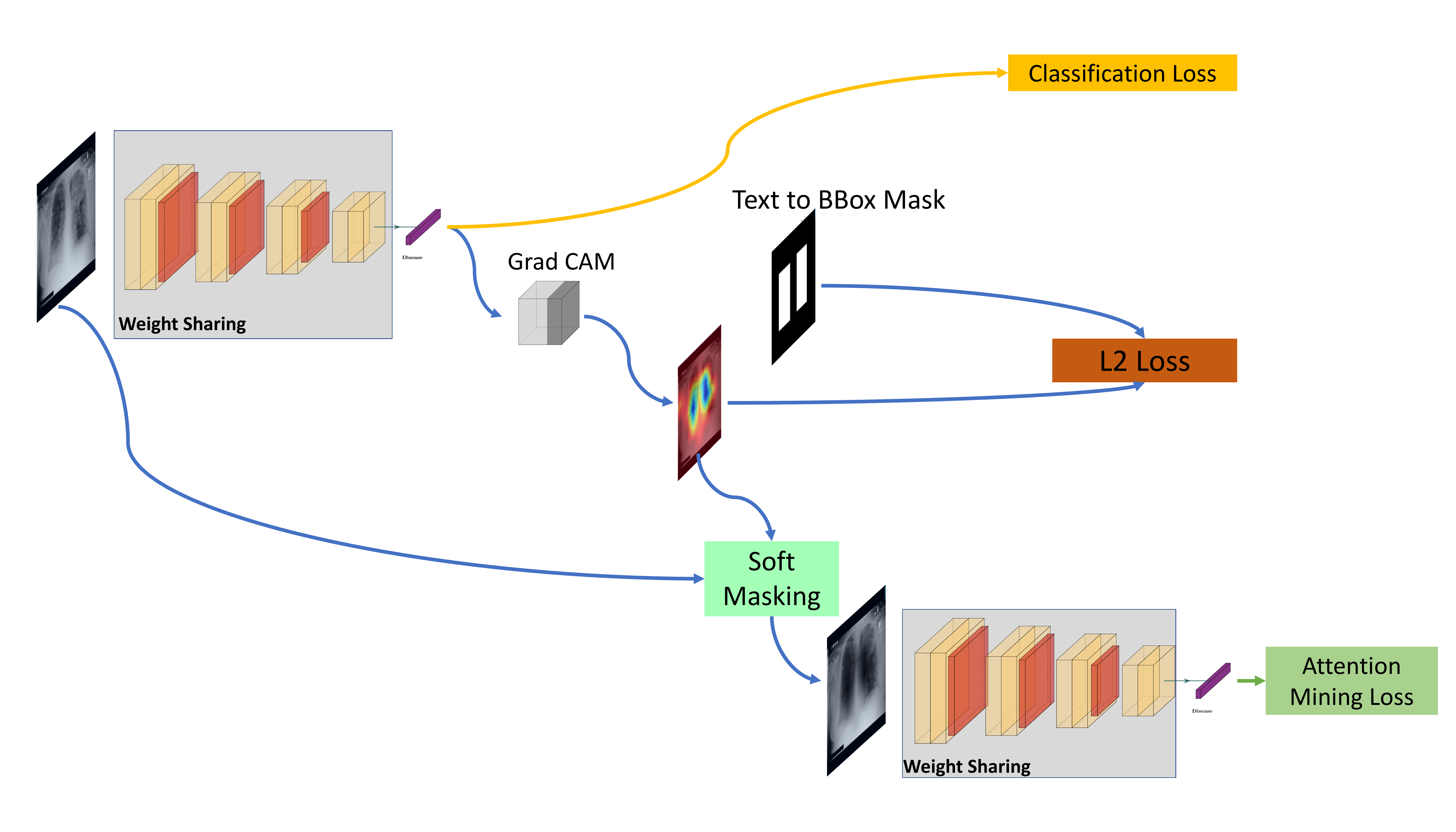}
		\caption{Flowchart of loss calculation: Starting from left, a chest X-ray is fed into the network and its classification loss (yellow) is calculated. Afterwards, using Grad CAM, the activation map is extracted. Using the bounding box mask from Stage 1 and the activation map the pixel wise L2 loss (orange) is calculated. Finally, the same activation map is used to soft mask the original chest X-ray image. The soft-masked image is fed again into the same network and its attention mining loss (green) is calculated. Note the network weights are shared between between all the loss calculations. Image modified from source:  \cite{DBLP:journals/corr/abs-1802-10171}}
		\label{fig:gainBlock}
	\end{centering}
\end{figure}

The block diagram of the proposed deep learning based system is shown in Figs. \ref{fig:texttobbox}, \ref{fig:gainBlock}. The algorithm can be divided into two parts:  a. The text to bounding box system, b. The Guided Attention based Inference Network (GAIN) \cite{DBLP:journals/corr/abs-1802-10171} which uses the pixel level bounding box labels to improve the image level classification. We used left and right opacity detection as the classification problem. 

\subsection{Text to Bounding Box System}

In order to provide an estimate of the diseased regions in the image,  we introduce a method of creating approximate bounding boxes from structured location modifiers. Table \ref{table:loclabels} shows the set of labels that were used to describe the bounding box locations in the chest radiographs. 


\begin{table}
	\caption{Table shows the list of location modifiers chosen by expert radiologists to describe the bounding box locations on a chest radiograph.}
	\begin{adjustbox}{width={\columnwidth},keepaspectratio}
		\begin{tabular}{l}
			\hline
			\multicolumn{1}{|l|}{right upper lung zone, right mid lung zone, right hilar structures, left upper lung zone}             \\ \hline
			\multicolumn{1}{|l|}{left mid lung zone, right hemidiaphragm, right lower lung zone, left hemidiaphragm}         \\ \hline
			\multicolumn{1}{|l|}{left lower lung zone, left hilar structures, right cardiophrenic angle}                       \\ \hline
			
			\multicolumn{1}{|l|}{right cardiac silhouette, left cardiophrenic angle, left cardiac silhouette}               \\ \hline
			\multicolumn{1}{|l|}{left costophrenic angle, right costophrenic angle, upper mediastinum}                          \\ \hline
			\label{table:loclabels}
		\end{tabular}
	\end{adjustbox}
\end{table}

\begin{figure}[htbp]
	\begin{centering}
		\includegraphics[width=\columnwidth]{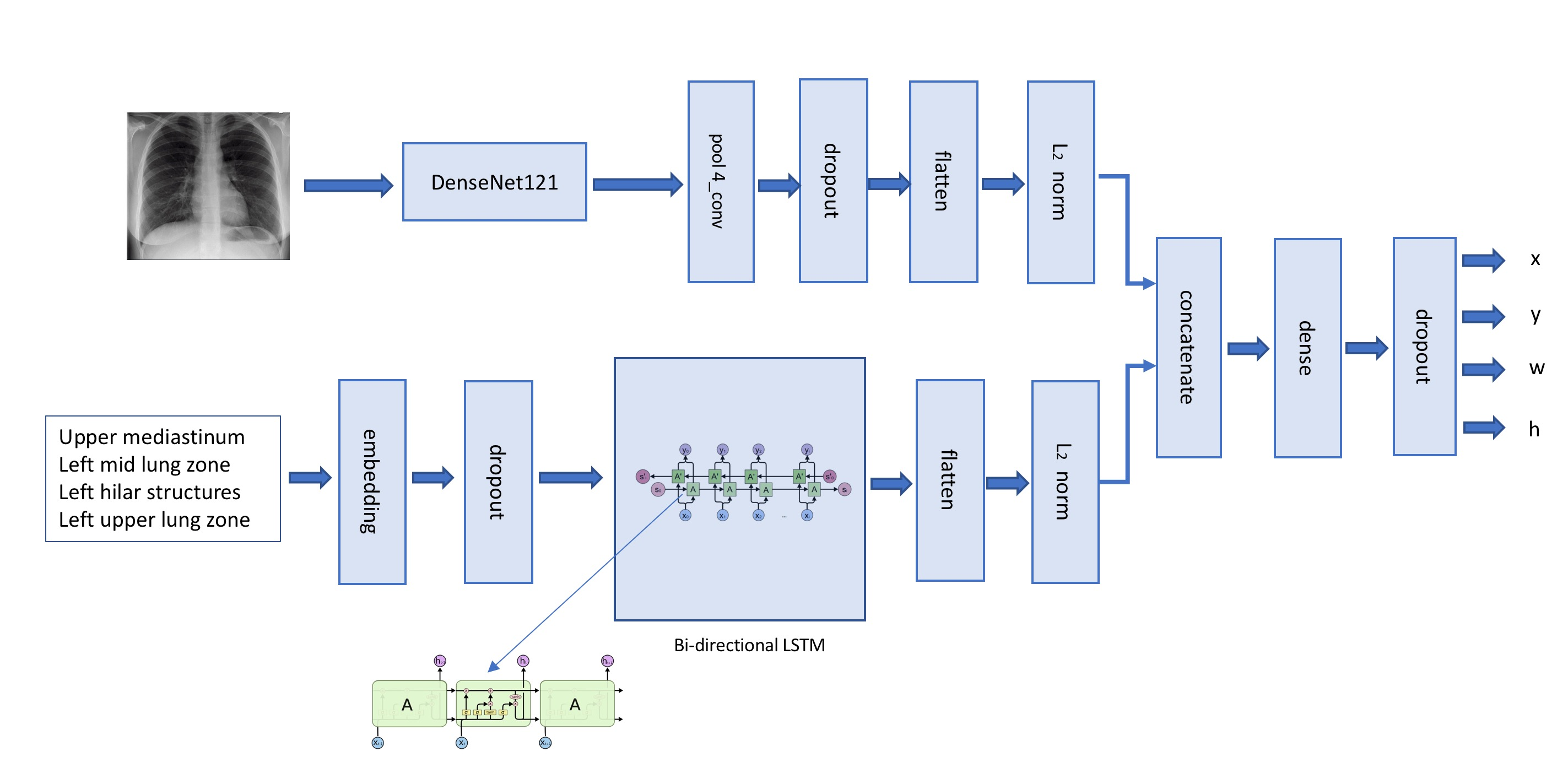}
		\caption{Text to bounding box network. Image modified from source: \url{http://colah.github.io/posts/2015-08-Understanding-LSTMs/}}
		\label{fig:texttobbox}
	\end{centering}
\end{figure}

The network designed was a combination of a pre-trained DenseNet-121 and a Bi-LSTM network. The two network outputs were combined to predict the bounding boxes. The hybrid network was trained end-to end with the inputs being the radiographs and the text based location modifiers to detect the bounding boxes of the image. The outputs of the network were the bounding box parameters (x, y, width and height) normalized between 0 to 1.  The overall network architecture design is shown in Fig. \ref{fig:texttobbox}. The DenseNet was pre-trained using ImageNet and used as a feature vector generator for the overall network. The outputs of the `pool 4 conv` layer was used as the output feature vector followed by an average pooling and Dense layers to get the vectorized output.

For processing the text-based location modifiers, the Bi-LSTM network was used where in the information flow in the their neurons split into the forward and the backward states. 



The training and the validation dataset for this method were based on the bounding boxes provided for the RSNA 2018 Pneumonia Detection Challenge  \footnote{https://www.kaggle.com/c/rsna-pneumonia-detection-challenge}. The radiologists looked at the bounding boxes per image and annotated them with text based locations from Table \ref{table:loclabels} as descriptors. 

Within the end to end classification algorithm, this text to image mapping network was used to produce bounding box localization cues. The disease classifier was trained on the MIMIC dataset \cite{DBLP:journals/corr/abs-1901-07042}. To extract the localization labels for this dataset, we relied on an Natural language processing (NLP) pipeline. 

\subsection{NLP for local label extraction}
We utilized a dictionary driven rule-based NLP algorithm to extract labels from CXR reports. A panel of 6 radiologists determined the key abnormal findings in the lungs and mediastinum at the granularity of children labels specified in Table \ref{table:nlp}. Two radiologists utilized a web-based unsupervised concept expansion NLP tool to efficiently expand the dictionary for each children label. The NLP tool proposes candidate terms from a corpus of ~500,000 raw CXR reports, and the radiologists validated any relevant abbreviations, misspelling, synonyms and different semantic ways of describing or implying the same label concept \cite{Coden2012}. This step reduced the number of non-meaningfully different target labels from reports for training imaging algorithms. The NLP algorithm first identifies the mention of a finding label in a report then uses a set of rules to determine if the finding was mentioned in the positive, negative or hypothetical context. For positive labels, the relevant sentence is searched for laterality and anatomical location cues. If no location information is mentioned in the sentence, a set of clinically driven rules determines the default locations that is most likely based on disease processes for the type of finding. The same concept expansion tool was used to expand the dictionary of terms for the different laterality and anatomical location labels in CXRs shown in Table \ref{table:nlp}. Finally, the children labels along with their laterality and location information is rolled up to the parent labels.

\begin{table}[htbp]
	\caption{Hierarchy of finding labels and associated relationship with laterality and location labels}
	\includegraphics[width=\columnwidth]{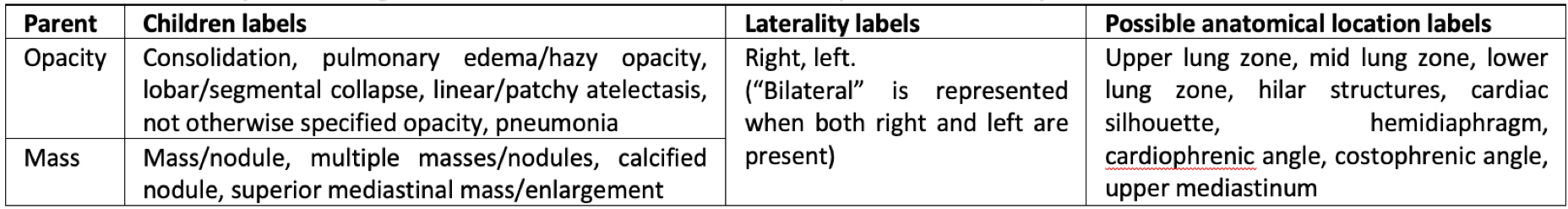}
	\label{table:nlp}
\end{table}

\subsection{Image Classification with NLP Parsed Bounding Boxes}

In part 2 of our system, we use the GAIN network that was proposed by \cite{DBLP:journals/corr/abs-1802-10171}. Their paper introduces a flexible methodology where they focus on interpretability of the network by improving the attention regions of the deep learning network. They demonstrate that by forcing the network to improve the attention maps of the outputs, they are able to improve the overall accuracy on the natural images. The outline of the algorithm is shown in Fig. \ref{fig:gainBlock}. The attention of the network was improved by proposing two additional losses called the attention mining loss and the external loss. To compute the attention mining loss, the network's final convolution layer was used to compute a Grad-CAM \cite{gradcam2016} on the truth labels. This attention map is masked onto the original image and the masked image is passed through the network again. The output probabilities of the computed image is used to compute the attention loss. The intuitive explanation of this loss is that whenever there are some other regions of the attention that get highlighted and provide a high probability for the truth label, the network is penalized to learn the full coverage of the attention region. Further, to improve the localization and interpretation of the network, a flexible module is presented where in the pixel level, the error between the attention maps and the masks from the text to bounding box system is computed. The external loss corresponds to the pixel wise loss between the ground truth image and the attention image. The overall loss function is thus formulated as : 

\begin{equation}
L = \underbrace{\lambda L_c}_\text{Classification Loss} + \overbrace{\alpha \frac{1}{n} \sum_c s^c(I^{*c})}^\text{Attention Mining Loss} + \underbrace{\omega \frac{1}{n}\sum_c(A^c - H^c)^2}_\text{Pixel Loss} 
\end{equation}

were $\lambda,\  \alpha,$\  and $ \omega$ are hyper parameters heuristically determined based on the application. $L_c$ is the standard classification loss while the second term is the probablity on the truth class for the masked image followed by the $\ell_2$ norm between the ground truth images and the attention image. The pixel level loss in an optional parameter which can be used only when the pixel level annotations were available. If not, the image level annotations could still be used demonstrating the flexibility of the system. 

\subsection{Experimental Methods}

The dataset for the text to bounding box system was a subset of the RSNA Pneumonia Dataset. The system was trained on 2000 images with bounding boxes and validated on a set of 643 images. The Bi-LSTM network was trained with 256 time steps with a dropout=0.25 and recurrent dropout=0.1.

To validate the GAIN network a combination the MIMIC dataset and a subset of NIH dataset for which our in-house radiologists have generated reports were used. Using NLP analysis, we isolated samples in these datasets with very strong indication of opacity and the laterality of this finding. The dataset was split into train, test and validation consisted of 57430, 8074, and 16071 respectively with fairly balanced distribution of left and right opacity images. We train the network to provide the right opacity or left opacity labels. The base network consisted of a ResNet-101 pretrained using ImageNet. The parameters used were:  Epochs=20, learning rate = 0.01, batch size =32, binary cross entropy loss with Adam optimizer. 

For the GAIN network, the baseline ResNet was fine tuned using the following parameters: Epochs = 50, learning rate = 0.001, $\lambda$ = 5.0, $\alpha$ = 1.0, $\omega$ = 10.0. The classification loss $L_c$ was a binary cross entropy loss with Adam optimizer. 
\begin{figure}
	\begin{centering}
		\includegraphics[width=\columnwidth]{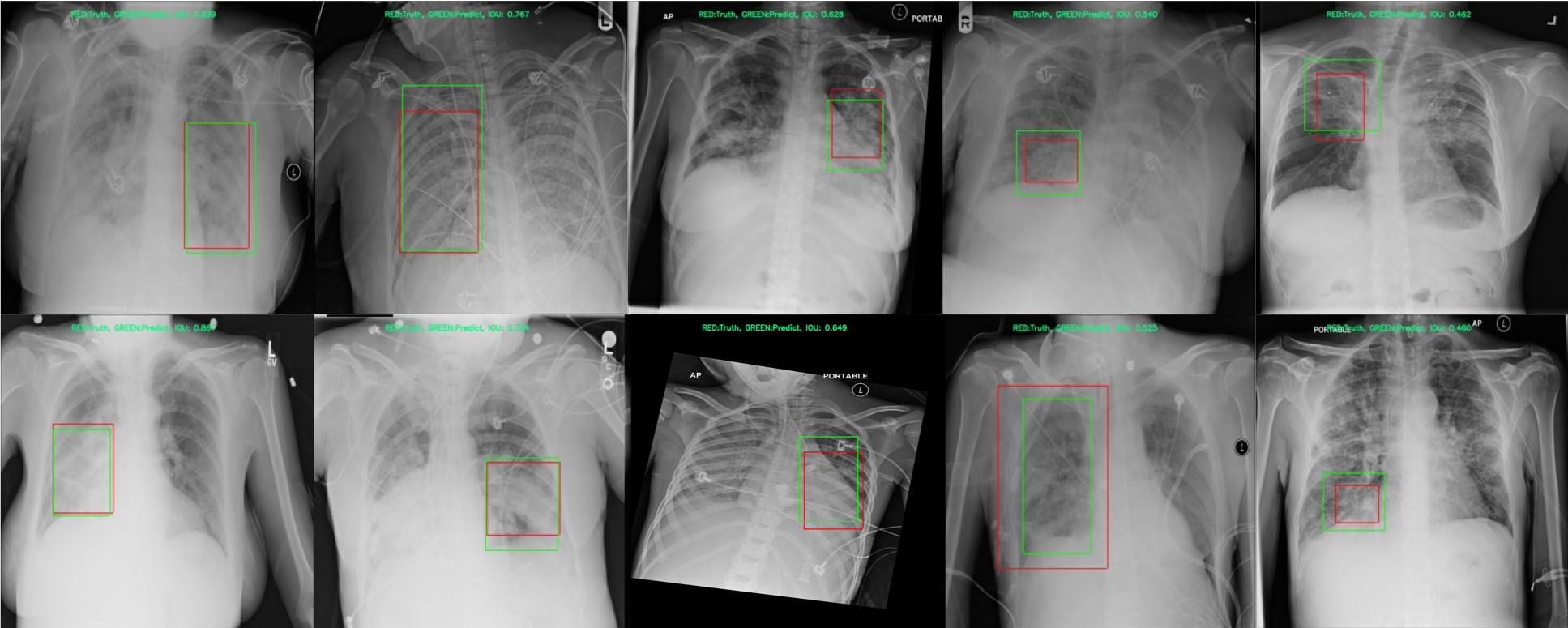}
		\caption{Figure shows some example bounding box outputs predicted by the network on the validation dataset. The red outlined box is the ground truth annotated by clinicians while the  green outlined box is the network prediction.}
		\label{fig:miou}
	\end{centering}
\end{figure}

\section{Results}
\subsection{Text to Bounding Box Results:} The map Intersection over Union (mIOU) was used to validate the text to bounding box performance. The average mIOU over the validation dataset from the RSNA challenge was \textbf{0.544}. Fig. \ref{fig:miou} shows some example bounding box overlaps between the ground truth and the predicted bounding boxes. Generally, the bounding boxes were well captured and indicated the approximate location of the region. 



\subsection{Classification and Attention Map Results:} 
\begin{figure}[htp]
	\includegraphics[width=\columnwidth]{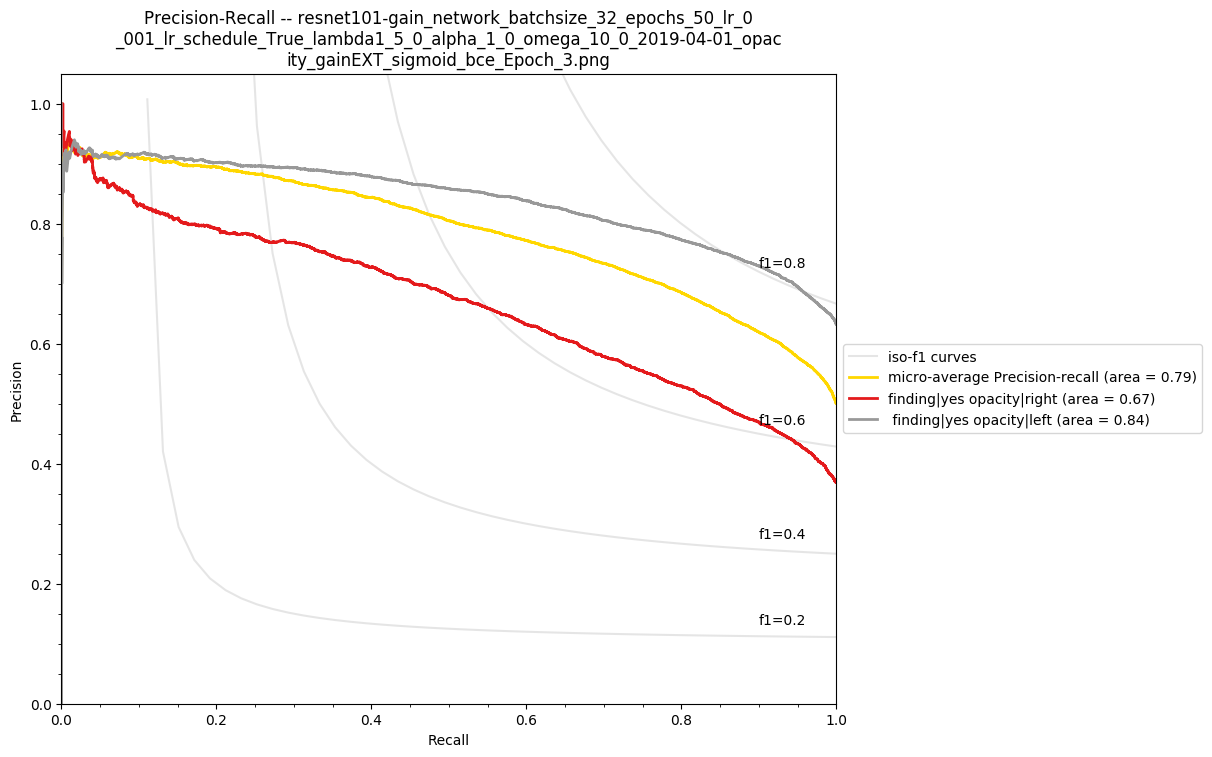}
	\includegraphics[width=\columnwidth]{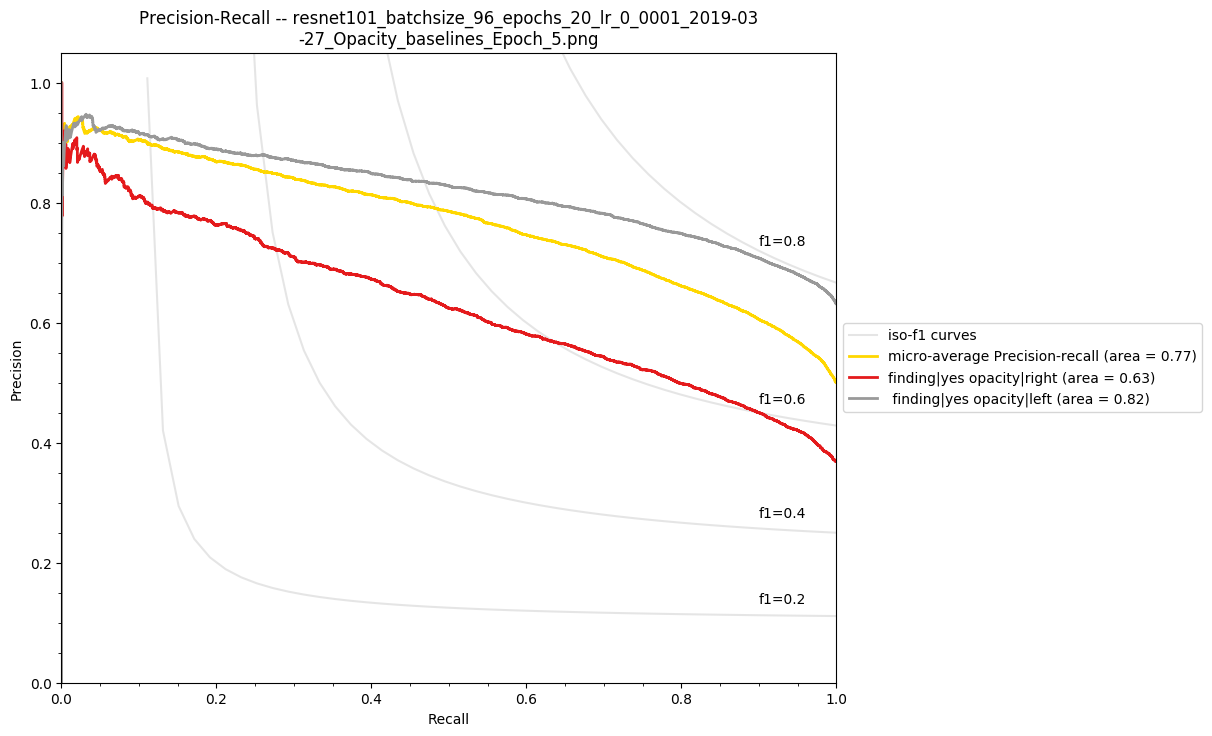}
	\includegraphics[width=\columnwidth]{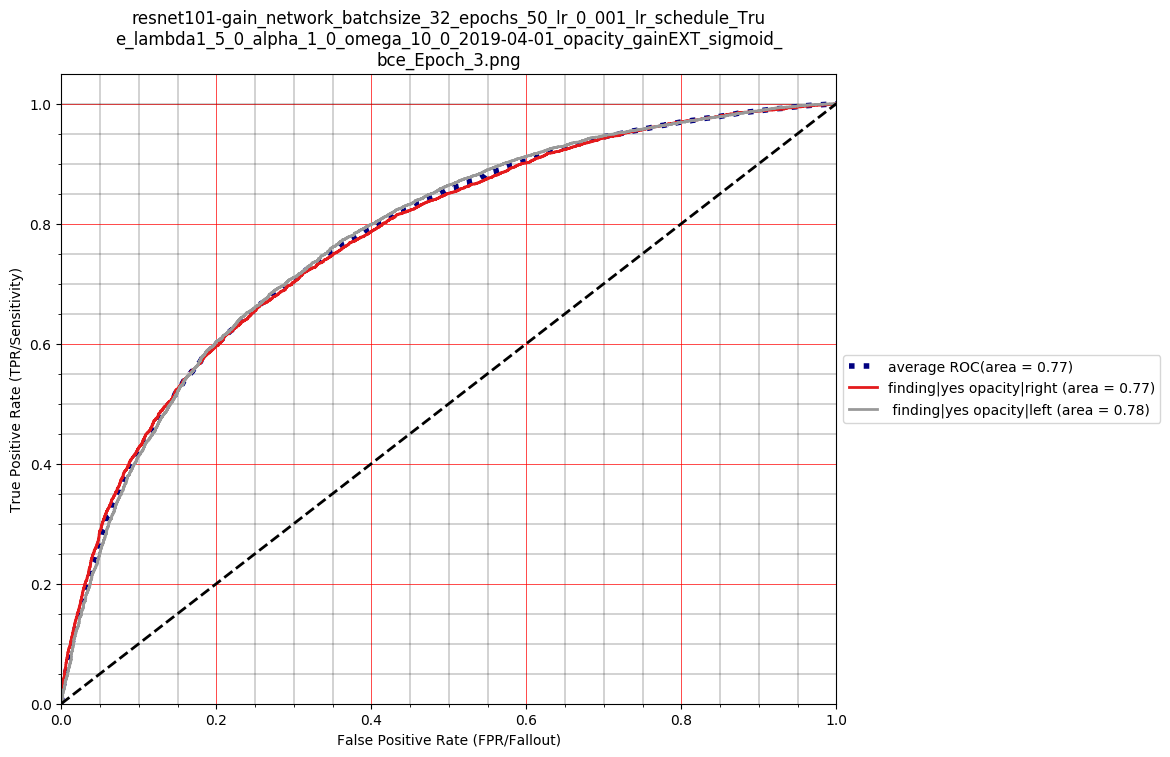}
	\includegraphics[width=\columnwidth]{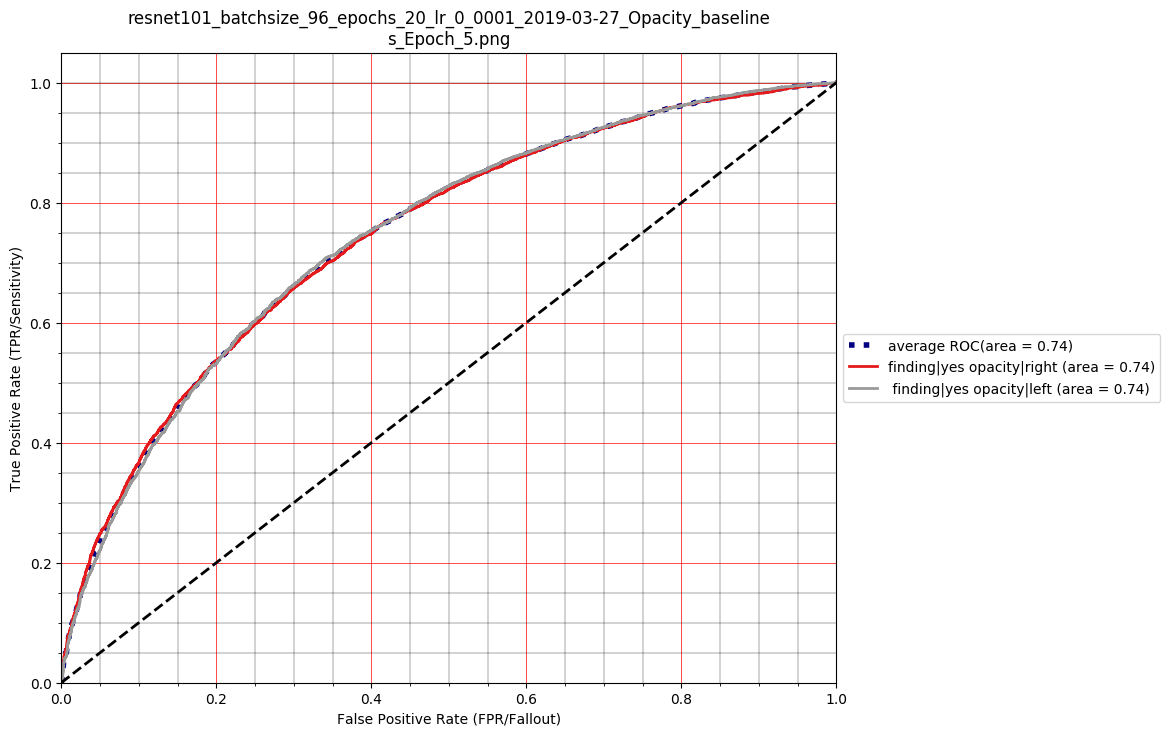}
	\caption{ AUROC for the case of left and right opacity detection were computed for the GAIN network and the baseline ResNet-101. The baseline area under PR curve = 0.63, 0.82 and AUROC = 0.74, 0.74  for right and left opacity respectively. Similarly for the GAIN network the PR curve = \textbf{0.67, 0.84} and AUROC = \textbf{0.77, 0.78} for right and left opacity respectively.}
	\label{fig:pr-curves}
\end{figure}

\begin{figure}
	\begin{centering}
		\includegraphics[width=\columnwidth]{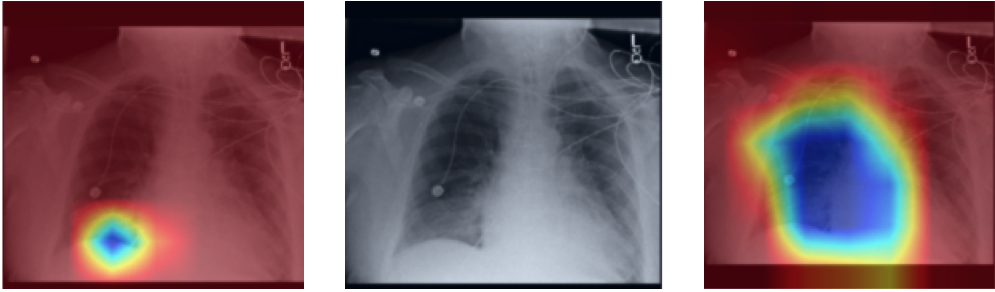}
		\includegraphics[width=\columnwidth]{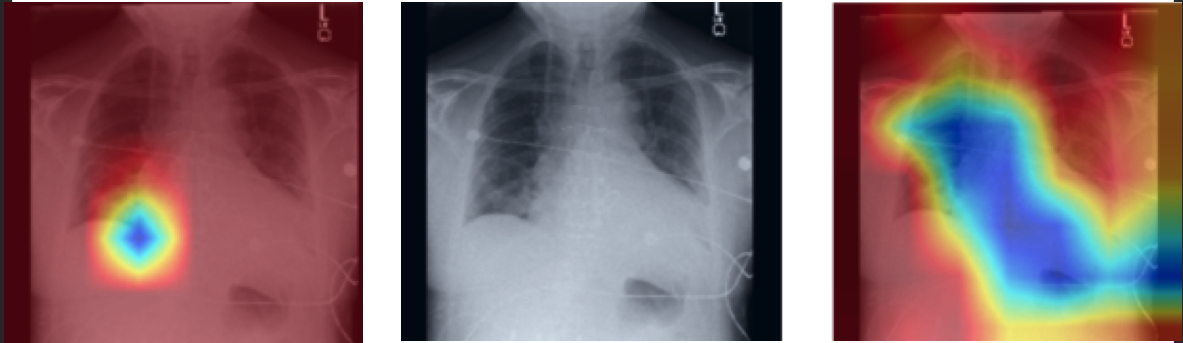}
		\includegraphics[width=\columnwidth]{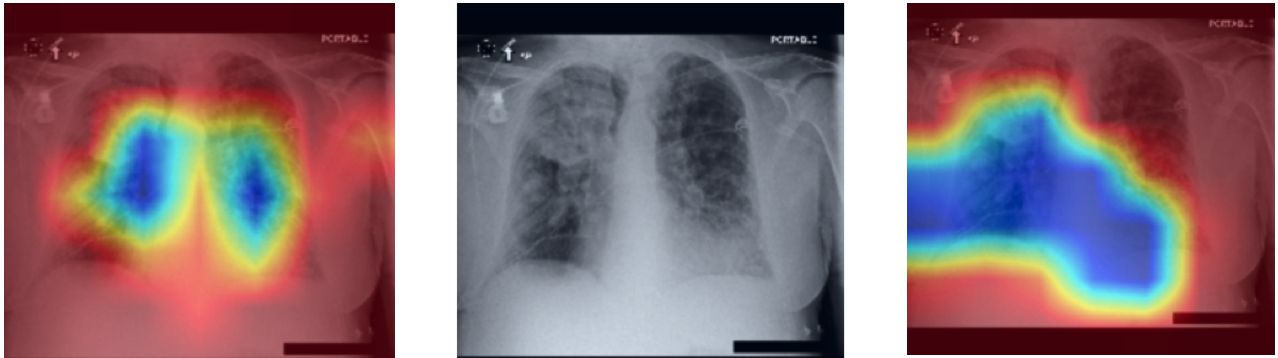}		
		\caption{Comparison of activation maps between a baseline (right) Resnet-101 and the Resnet-101 trained using the GAIN method (left). Left, centre and right columns shows the GAIN activation, original image and the baseline.}
		\label{fig:activation}
	\end{centering}
\end{figure}

The precision recall (PR) and the area under the receiver operation curve (AUROC) for the case of left and right opacity detection were computed for the GAIN network and the baseline ResNet-101. The baseline area under PR curve = 0.63, 0.82 and AUROC = 0.74, 0.74  for right and left opacity respectively. Similarly for the GAIN network the PR curve = \textbf{0.67, 0.84} and AUROC = \textbf{0.77, 0.78} for right and left opacity respectively. The ROC and PR curves can be see in Fig. \ref{fig:pr-curves}. Further we analyze the activation maps generated for a few example cases of the baseline and the GAIN network results. We clearly see that in Fig. \ref{fig:activation}, the GAIN network is able to focus on the regions of opacity thereby improving the accuracy and the interpretability of the system. 

\section{Discussions and Conclusion}

Apart from the value that this work provides for improving classification results, it can also have a practical value. The current practice of radiology consists of visual inspection followed by dictating a text report. The proposed method here provides the possibility of enhancing this approach through generating a marking on the image linked to a description in the text. The attention map can be the starting point of this marked region. Inclusion of local annotations with text reports can expedite follow up reviews and improve the radiology work-flow.

We showed that by extracting the expected anomaly location from automated text to bounding box network and biasing the subsequent attention-guided inference network (GAIN), one can achieve higher classification accuracy and focused attention maps for chest X-ray classification. The chosen problem of evaluating the spatial location of the opacity was particularly challenging especially when we observe the qualitative results of the baseline network as shown in Fig. \ref{fig:activation}. By using the proposed system, we observe that along with the accurate predictions of the opacity laterality, the spatial location of the disease is also captured. Further, we note that attention maps are a very qualitative measure of the networks output.

Although the original application of GAIN was more successful in terms of performance metrics, our benefits of using the GAIN network has been limited and we mostly saw improvement in the localization. This brings to question how correlated the localization of attention maps is with accuracy and remains part of our future studies. Further work will involve developing quantitative metrics to evaluate the attention maps more accurately. We currently work on expanding this pipeline to include a larger number of anomalies.



\bibliographystyle{IEEEbib}
\bibliography{Xray_Grand_Challenge-MICCAI_2019}

\end{document}